\pdfoutput=1

\documentclass[11pt]{article}

\usepackage{naacl2021}

\usepackage{times}
\usepackage{latexsym}

\usepackage{amsmath}
\usepackage{url}
\usepackage{amssymb}
\usepackage{amsfonts}
\usepackage{graphicx}
\usepackage{tabularx}
\usepackage{multirow}
\usepackage{arydshln}
\usepackage{mathtools,nccmath}
\usepackage{listings}

\usepackage[T5]{fontenc}
\usepackage{enumitem}
\usepackage{todonotes}
\usepackage{cancel}
\usepackage[draft]{minted}

\usepackage{microtype}

\makeatletter
\def\PYGdefault@reset{\let\PYGdefault@it=\relax \let\PYGdefault@bf=\relax%
    \let\PYGdefault@ul=\relax \let\PYGdefault@tc=\relax%
    \let\PYGdefault@bc=\relax \let\PYGdefault@ff=\relax}
\def\PYGdefault@tok#1{\csname PYGdefault@tok@#1\endcsname}
\def\PYGdefault@toks#1+{\ifx\relax#1\empty\else%
    \PYGdefault@tok{#1}\expandafter\PYGdefault@toks\fi}
\def\PYGdefault@do#1{\PYGdefault@bc{\PYGdefault@tc{\PYGdefault@ul{%
    \PYGdefault@it{\PYGdefault@bf{\PYGdefault@ff{#1}}}}}}}
\def\PYGdefault#1#2{\PYGdefault@reset\PYGdefault@toks#1+\relax+\PYGdefault@do{#2}}

\expandafter\def\csname PYGdefault@tok@w\endcsname{\def\PYGdefault@tc##1{\textcolor[rgb]{0.73,0.73,0.73}{##1}}}
\expandafter\def\csname PYGdefault@tok@c\endcsname{\let\PYGdefault@it=\textit\def\PYGdefault@tc##1{\textcolor[rgb]{0.25,0.50,0.50}{##1}}}
\expandafter\def\csname PYGdefault@tok@cp\endcsname{\def\PYGdefault@tc##1{\textcolor[rgb]{0.74,0.48,0.00}{##1}}}
\expandafter\def\csname PYGdefault@tok@k\endcsname{\let\PYGdefault@bf=\textbf\def\PYGdefault@tc##1{\textcolor[rgb]{0.00,0.50,0.00}{##1}}}
\expandafter\def\csname PYGdefault@tok@kp\endcsname{\def\PYGdefault@tc##1{\textcolor[rgb]{0.00,0.50,0.00}{##1}}}
\expandafter\def\csname PYGdefault@tok@kt\endcsname{\def\PYGdefault@tc##1{\textcolor[rgb]{0.69,0.00,0.25}{##1}}}
\expandafter\def\csname PYGdefault@tok@o\endcsname{\def\PYGdefault@tc##1{\textcolor[rgb]{0.40,0.40,0.40}{##1}}}
\expandafter\def\csname PYGdefault@tok@ow\endcsname{\let\PYGdefault@bf=\textbf\def\PYGdefault@tc##1{\textcolor[rgb]{0.67,0.13,1.00}{##1}}}
\expandafter\def\csname PYGdefault@tok@nb\endcsname{\def\PYGdefault@tc##1{\textcolor[rgb]{0.00,0.50,0.00}{##1}}}
\expandafter\def\csname PYGdefault@tok@nf\endcsname{\def\PYGdefault@tc##1{\textcolor[rgb]{0.00,0.00,1.00}{##1}}}
\expandafter\def\csname PYGdefault@tok@nc\endcsname{\let\PYGdefault@bf=\textbf\def\PYGdefault@tc##1{\textcolor[rgb]{0.00,0.00,1.00}{##1}}}
\expandafter\def\csname PYGdefault@tok@nn\endcsname{\let\PYGdefault@bf=\textbf\def\PYGdefault@tc##1{\textcolor[rgb]{0.00,0.00,1.00}{##1}}}
\expandafter\def\csname PYGdefault@tok@ne\endcsname{\let\PYGdefault@bf=\textbf\def\PYGdefault@tc##1{\textcolor[rgb]{0.82,0.25,0.23}{##1}}}
\expandafter\def\csname PYGdefault@tok@nv\endcsname{\def\PYGdefault@tc##1{\textcolor[rgb]{0.10,0.09,0.49}{##1}}}
\expandafter\def\csname PYGdefault@tok@no\endcsname{\def\PYGdefault@tc##1{\textcolor[rgb]{0.53,0.00,0.00}{##1}}}
\expandafter\def\csname PYGdefault@tok@nl\endcsname{\def\PYGdefault@tc##1{\textcolor[rgb]{0.63,0.63,0.00}{##1}}}
\expandafter\def\csname PYGdefault@tok@ni\endcsname{\let\PYGdefault@bf=\textbf\def\PYGdefault@tc##1{\textcolor[rgb]{0.60,0.60,0.60}{##1}}}
\expandafter\def\csname PYGdefault@tok@na\endcsname{\def\PYGdefault@tc##1{\textcolor[rgb]{0.49,0.56,0.16}{##1}}}
\expandafter\def\csname PYGdefault@tok@nt\endcsname{\let\PYGdefault@bf=\textbf\def\PYGdefault@tc##1{\textcolor[rgb]{0.00,0.50,0.00}{##1}}}
\expandafter\def\csname PYGdefault@tok@nd\endcsname{\def\PYGdefault@tc##1{\textcolor[rgb]{0.67,0.13,1.00}{##1}}}
\expandafter\def\csname PYGdefault@tok@s\endcsname{\def\PYGdefault@tc##1{\textcolor[rgb]{0.73,0.13,0.13}{##1}}}
\expandafter\def\csname PYGdefault@tok@sd\endcsname{\let\PYGdefault@it=\textit\def\PYGdefault@tc##1{\textcolor[rgb]{0.73,0.13,0.13}{##1}}}
\expandafter\def\csname PYGdefault@tok@si\endcsname{\let\PYGdefault@bf=\textbf\def\PYGdefault@tc##1{\textcolor[rgb]{0.73,0.40,0.53}{##1}}}
\expandafter\def\csname PYGdefault@tok@se\endcsname{\let\PYGdefault@bf=\textbf\def\PYGdefault@tc##1{\textcolor[rgb]{0.73,0.40,0.13}{##1}}}
\expandafter\def\csname PYGdefault@tok@sr\endcsname{\def\PYGdefault@tc##1{\textcolor[rgb]{0.73,0.40,0.53}{##1}}}
\expandafter\def\csname PYGdefault@tok@ss\endcsname{\def\PYGdefault@tc##1{\textcolor[rgb]{0.10,0.09,0.49}{##1}}}
\expandafter\def\csname PYGdefault@tok@sx\endcsname{\def\PYGdefault@tc##1{\textcolor[rgb]{0.00,0.50,0.00}{##1}}}
\expandafter\def\csname PYGdefault@tok@m\endcsname{\def\PYGdefault@tc##1{\textcolor[rgb]{0.40,0.40,0.40}{##1}}}
\expandafter\def\csname PYGdefault@tok@gh\endcsname{\let\PYGdefault@bf=\textbf\def\PYGdefault@tc##1{\textcolor[rgb]{0.00,0.00,0.50}{##1}}}
\expandafter\def\csname PYGdefault@tok@gu\endcsname{\let\PYGdefault@bf=\textbf\def\PYGdefault@tc##1{\textcolor[rgb]{0.50,0.00,0.50}{##1}}}
\expandafter\def\csname PYGdefault@tok@gd\endcsname{\def\PYGdefault@tc##1{\textcolor[rgb]{0.63,0.00,0.00}{##1}}}
\expandafter\def\csname PYGdefault@tok@gi\endcsname{\def\PYGdefault@tc##1{\textcolor[rgb]{0.00,0.63,0.00}{##1}}}
\expandafter\def\csname PYGdefault@tok@gr\endcsname{\def\PYGdefault@tc##1{\textcolor[rgb]{1.00,0.00,0.00}{##1}}}
\expandafter\def\csname PYGdefault@tok@ge\endcsname{\let\PYGdefault@it=\textit}
\expandafter\def\csname PYGdefault@tok@gs\endcsname{\let\PYGdefault@bf=\textbf}
\expandafter\def\csname PYGdefault@tok@gp\endcsname{\let\PYGdefault@bf=\textbf\def\PYGdefault@tc##1{\textcolor[rgb]{0.00,0.00,0.50}{##1}}}
\expandafter\def\csname PYGdefault@tok@go\endcsname{\def\PYGdefault@tc##1{\textcolor[rgb]{0.53,0.53,0.53}{##1}}}
\expandafter\def\csname PYGdefault@tok@gt\endcsname{\def\PYGdefault@tc##1{\textcolor[rgb]{0.00,0.27,0.87}{##1}}}
\expandafter\def\csname PYGdefault@tok@err\endcsname{\def\PYGdefault@bc##1{\setlength{\fboxsep}{0pt}\fcolorbox[rgb]{1.00,0.00,0.00}{1,1,1}{\strut ##1}}}
\expandafter\def\csname PYGdefault@tok@kc\endcsname{\let\PYGdefault@bf=\textbf\def\PYGdefault@tc##1{\textcolor[rgb]{0.00,0.50,0.00}{##1}}}
\expandafter\def\csname PYGdefault@tok@kd\endcsname{\let\PYGdefault@bf=\textbf\def\PYGdefault@tc##1{\textcolor[rgb]{0.00,0.50,0.00}{##1}}}
\expandafter\def\csname PYGdefault@tok@kn\endcsname{\let\PYGdefault@bf=\textbf\def\PYGdefault@tc##1{\textcolor[rgb]{0.00,0.50,0.00}{##1}}}
\expandafter\def\csname PYGdefault@tok@kr\endcsname{\let\PYGdefault@bf=\textbf\def\PYGdefault@tc##1{\textcolor[rgb]{0.00,0.50,0.00}{##1}}}
\expandafter\def\csname PYGdefault@tok@bp\endcsname{\def\PYGdefault@tc##1{\textcolor[rgb]{0.00,0.50,0.00}{##1}}}
\expandafter\def\csname PYGdefault@tok@fm\endcsname{\def\PYGdefault@tc##1{\textcolor[rgb]{0.00,0.00,1.00}{##1}}}
\expandafter\def\csname PYGdefault@tok@vc\endcsname{\def\PYGdefault@tc##1{\textcolor[rgb]{0.10,0.09,0.49}{##1}}}
\expandafter\def\csname PYGdefault@tok@vg\endcsname{\def\PYGdefault@tc##1{\textcolor[rgb]{0.10,0.09,0.49}{##1}}}
\expandafter\def\csname PYGdefault@tok@vi\endcsname{\def\PYGdefault@tc##1{\textcolor[rgb]{0.10,0.09,0.49}{##1}}}
\expandafter\def\csname PYGdefault@tok@vm\endcsname{\def\PYGdefault@tc##1{\textcolor[rgb]{0.10,0.09,0.49}{##1}}}
\expandafter\def\csname PYGdefault@tok@sa\endcsname{\def\PYGdefault@tc##1{\textcolor[rgb]{0.73,0.13,0.13}{##1}}}
\expandafter\def\csname PYGdefault@tok@sb\endcsname{\def\PYGdefault@tc##1{\textcolor[rgb]{0.73,0.13,0.13}{##1}}}
\expandafter\def\csname PYGdefault@tok@sc\endcsname{\def\PYGdefault@tc##1{\textcolor[rgb]{0.73,0.13,0.13}{##1}}}
\expandafter\def\csname PYGdefault@tok@dl\endcsname{\def\PYGdefault@tc##1{\textcolor[rgb]{0.73,0.13,0.13}{##1}}}
\expandafter\def\csname PYGdefault@tok@s2\endcsname{\def\PYGdefault@tc##1{\textcolor[rgb]{0.73,0.13,0.13}{##1}}}
\expandafter\def\csname PYGdefault@tok@sh\endcsname{\def\PYGdefault@tc##1{\textcolor[rgb]{0.73,0.13,0.13}{##1}}}
\expandafter\def\csname PYGdefault@tok@s1\endcsname{\def\PYGdefault@tc##1{\textcolor[rgb]{0.73,0.13,0.13}{##1}}}
\expandafter\def\csname PYGdefault@tok@mb\endcsname{\def\PYGdefault@tc##1{\textcolor[rgb]{0.40,0.40,0.40}{##1}}}
\expandafter\def\csname PYGdefault@tok@mf\endcsname{\def\PYGdefault@tc##1{\textcolor[rgb]{0.40,0.40,0.40}{##1}}}
\expandafter\def\csname PYGdefault@tok@mh\endcsname{\def\PYGdefault@tc##1{\textcolor[rgb]{0.40,0.40,0.40}{##1}}}
\expandafter\def\csname PYGdefault@tok@mi\endcsname{\def\PYGdefault@tc##1{\textcolor[rgb]{0.40,0.40,0.40}{##1}}}
\expandafter\def\csname PYGdefault@tok@il\endcsname{\def\PYGdefault@tc##1{\textcolor[rgb]{0.40,0.40,0.40}{##1}}}
\expandafter\def\csname PYGdefault@tok@mo\endcsname{\def\PYGdefault@tc##1{\textcolor[rgb]{0.40,0.40,0.40}{##1}}}
\expandafter\def\csname PYGdefault@tok@ch\endcsname{\let\PYGdefault@it=\textit\def\PYGdefault@tc##1{\textcolor[rgb]{0.25,0.50,0.50}{##1}}}
\expandafter\def\csname PYGdefault@tok@cm\endcsname{\let\PYGdefault@it=\textit\def\PYGdefault@tc##1{\textcolor[rgb]{0.25,0.50,0.50}{##1}}}
\expandafter\def\csname PYGdefault@tok@cpf\endcsname{\let\PYGdefault@it=\textit\def\PYGdefault@tc##1{\textcolor[rgb]{0.25,0.50,0.50}{##1}}}
\expandafter\def\csname PYGdefault@tok@c1\endcsname{\let\PYGdefault@it=\textit\def\PYGdefault@tc##1{\textcolor[rgb]{0.25,0.50,0.50}{##1}}}
\expandafter\def\csname PYGdefault@tok@cs\endcsname{\let\PYGdefault@it=\textit\def\PYGdefault@tc##1{\textcolor[rgb]{0.25,0.50,0.50}{##1}}}

\makeatother

\makeatletter
\def\PYG@reset{\let\PYG@it=\relax \let\PYG@bf=\relax%
    \let\PYG@ul=\relax \let\PYG@tc=\relax%
    \let\PYG@bc=\relax \let\PYG@ff=\relax}
\def\PYG@tok#1{\csname PYG@tok@#1\endcsname}
\def\PYG@toks#1+{\ifx\relax#1\empty\else%
    \PYG@tok{#1}\expandafter\PYG@toks\fi}
\def\PYG@do#1{\PYG@bc{\PYG@tc{\PYG@ul{%
    \PYG@it{\PYG@bf{\PYG@ff{#1}}}}}}}
\def\PYG#1#2{\PYG@reset\PYG@toks#1+\relax+\PYG@do{#2}}

\expandafter\def\csname PYG@tok@w\endcsname{\def\PYG@tc##1{\textcolor[rgb]{0.73,0.73,0.73}{##1}}}
\expandafter\def\csname PYG@tok@c\endcsname{\let\PYG@it=\textit\def\PYG@tc##1{\textcolor[rgb]{0.25,0.50,0.50}{##1}}}
\expandafter\def\csname PYG@tok@cp\endcsname{\def\PYG@tc##1{\textcolor[rgb]{0.74,0.48,0.00}{##1}}}
\expandafter\def\csname PYG@tok@k\endcsname{\let\PYG@bf=\textbf\def\PYG@tc##1{\textcolor[rgb]{0.00,0.50,0.00}{##1}}}
\expandafter\def\csname PYG@tok@kp\endcsname{\def\PYG@tc##1{\textcolor[rgb]{0.00,0.50,0.00}{##1}}}
\expandafter\def\csname PYG@tok@kt\endcsname{\def\PYG@tc##1{\textcolor[rgb]{0.69,0.00,0.25}{##1}}}
\expandafter\def\csname PYG@tok@o\endcsname{\def\PYG@tc##1{\textcolor[rgb]{0.40,0.40,0.40}{##1}}}
\expandafter\def\csname PYG@tok@ow\endcsname{\let\PYG@bf=\textbf\def\PYG@tc##1{\textcolor[rgb]{0.67,0.13,1.00}{##1}}}
\expandafter\def\csname PYG@tok@nb\endcsname{\def\PYG@tc##1{\textcolor[rgb]{0.00,0.50,0.00}{##1}}}
\expandafter\def\csname PYG@tok@nf\endcsname{\def\PYG@tc##1{\textcolor[rgb]{0.00,0.00,1.00}{##1}}}
\expandafter\def\csname PYG@tok@nc\endcsname{\let\PYG@bf=\textbf\def\PYG@tc##1{\textcolor[rgb]{0.00,0.00,1.00}{##1}}}
\expandafter\def\csname PYG@tok@nn\endcsname{\let\PYG@bf=\textbf\def\PYG@tc##1{\textcolor[rgb]{0.00,0.00,1.00}{##1}}}
\expandafter\def\csname PYG@tok@ne\endcsname{\let\PYG@bf=\textbf\def\PYG@tc##1{\textcolor[rgb]{0.82,0.25,0.23}{##1}}}
\expandafter\def\csname PYG@tok@nv\endcsname{\def\PYG@tc##1{\textcolor[rgb]{0.10,0.09,0.49}{##1}}}
\expandafter\def\csname PYG@tok@no\endcsname{\def\PYG@tc##1{\textcolor[rgb]{0.53,0.00,0.00}{##1}}}
\expandafter\def\csname PYG@tok@nl\endcsname{\def\PYG@tc##1{\textcolor[rgb]{0.63,0.63,0.00}{##1}}}
\expandafter\def\csname PYG@tok@ni\endcsname{\let\PYG@bf=\textbf\def\PYG@tc##1{\textcolor[rgb]{0.60,0.60,0.60}{##1}}}
\expandafter\def\csname PYG@tok@na\endcsname{\def\PYG@tc##1{\textcolor[rgb]{0.49,0.56,0.16}{##1}}}
\expandafter\def\csname PYG@tok@nt\endcsname{\let\PYG@bf=\textbf\def\PYG@tc##1{\textcolor[rgb]{0.00,0.50,0.00}{##1}}}
\expandafter\def\csname PYG@tok@nd\endcsname{\def\PYG@tc##1{\textcolor[rgb]{0.67,0.13,1.00}{##1}}}
\expandafter\def\csname PYG@tok@s\endcsname{\def\PYG@tc##1{\textcolor[rgb]{0.73,0.13,0.13}{##1}}}
\expandafter\def\csname PYG@tok@sd\endcsname{\let\PYG@it=\textit\def\PYG@tc##1{\textcolor[rgb]{0.73,0.13,0.13}{##1}}}
\expandafter\def\csname PYG@tok@si\endcsname{\let\PYG@bf=\textbf\def\PYG@tc##1{\textcolor[rgb]{0.73,0.40,0.53}{##1}}}
\expandafter\def\csname PYG@tok@se\endcsname{\let\PYG@bf=\textbf\def\PYG@tc##1{\textcolor[rgb]{0.73,0.40,0.13}{##1}}}
\expandafter\def\csname PYG@tok@sr\endcsname{\def\PYG@tc##1{\textcolor[rgb]{0.73,0.40,0.53}{##1}}}
\expandafter\def\csname PYG@tok@ss\endcsname{\def\PYG@tc##1{\textcolor[rgb]{0.10,0.09,0.49}{##1}}}
\expandafter\def\csname PYG@tok@sx\endcsname{\def\PYG@tc##1{\textcolor[rgb]{0.00,0.50,0.00}{##1}}}
\expandafter\def\csname PYG@tok@m\endcsname{\def\PYG@tc##1{\textcolor[rgb]{0.40,0.40,0.40}{##1}}}
\expandafter\def\csname PYG@tok@gh\endcsname{\let\PYG@bf=\textbf\def\PYG@tc##1{\textcolor[rgb]{0.00,0.00,0.50}{##1}}}
\expandafter\def\csname PYG@tok@gu\endcsname{\let\PYG@bf=\textbf\def\PYG@tc##1{\textcolor[rgb]{0.50,0.00,0.50}{##1}}}
\expandafter\def\csname PYG@tok@gd\endcsname{\def\PYG@tc##1{\textcolor[rgb]{0.63,0.00,0.00}{##1}}}
\expandafter\def\csname PYG@tok@gi\endcsname{\def\PYG@tc##1{\textcolor[rgb]{0.00,0.63,0.00}{##1}}}
\expandafter\def\csname PYG@tok@gr\endcsname{\def\PYG@tc##1{\textcolor[rgb]{1.00,0.00,0.00}{##1}}}
\expandafter\def\csname PYG@tok@ge\endcsname{\let\PYG@it=\textit}
\expandafter\def\csname PYG@tok@gs\endcsname{\let\PYG@bf=\textbf}
\expandafter\def\csname PYG@tok@gp\endcsname{\let\PYG@bf=\textbf\def\PYG@tc##1{\textcolor[rgb]{0.00,0.00,0.50}{##1}}}
\expandafter\def\csname PYG@tok@go\endcsname{\def\PYG@tc##1{\textcolor[rgb]{0.53,0.53,0.53}{##1}}}
\expandafter\def\csname PYG@tok@gt\endcsname{\def\PYG@tc##1{\textcolor[rgb]{0.00,0.27,0.87}{##1}}}
\expandafter\def\csname PYG@tok@err\endcsname{\def\PYG@bc##1{\setlength{\fboxsep}{0pt}\fcolorbox[rgb]{1.00,0.00,0.00}{1,1,1}{\strut ##1}}}
\expandafter\def\csname PYG@tok@kc\endcsname{\let\PYG@bf=\textbf\def\PYG@tc##1{\textcolor[rgb]{0.00,0.50,0.00}{##1}}}
\expandafter\def\csname PYG@tok@kd\endcsname{\let\PYG@bf=\textbf\def\PYG@tc##1{\textcolor[rgb]{0.00,0.50,0.00}{##1}}}
\expandafter\def\csname PYG@tok@kn\endcsname{\let\PYG@bf=\textbf\def\PYG@tc##1{\textcolor[rgb]{0.00,0.50,0.00}{##1}}}
\expandafter\def\csname PYG@tok@kr\endcsname{\let\PYG@bf=\textbf\def\PYG@tc##1{\textcolor[rgb]{0.00,0.50,0.00}{##1}}}
\expandafter\def\csname PYG@tok@bp\endcsname{\def\PYG@tc##1{\textcolor[rgb]{0.00,0.50,0.00}{##1}}}
\expandafter\def\csname PYG@tok@fm\endcsname{\def\PYG@tc##1{\textcolor[rgb]{0.00,0.00,1.00}{##1}}}
\expandafter\def\csname PYG@tok@vc\endcsname{\def\PYG@tc##1{\textcolor[rgb]{0.10,0.09,0.49}{##1}}}
\expandafter\def\csname PYG@tok@vg\endcsname{\def\PYG@tc##1{\textcolor[rgb]{0.10,0.09,0.49}{##1}}}
\expandafter\def\csname PYG@tok@vi\endcsname{\def\PYG@tc##1{\textcolor[rgb]{0.10,0.09,0.49}{##1}}}
\expandafter\def\csname PYG@tok@vm\endcsname{\def\PYG@tc##1{\textcolor[rgb]{0.10,0.09,0.49}{##1}}}
\expandafter\def\csname PYG@tok@sa\endcsname{\def\PYG@tc##1{\textcolor[rgb]{0.73,0.13,0.13}{##1}}}
\expandafter\def\csname PYG@tok@sb\endcsname{\def\PYG@tc##1{\textcolor[rgb]{0.73,0.13,0.13}{##1}}}
\expandafter\def\csname PYG@tok@sc\endcsname{\def\PYG@tc##1{\textcolor[rgb]{0.73,0.13,0.13}{##1}}}
\expandafter\def\csname PYG@tok@dl\endcsname{\def\PYG@tc##1{\textcolor[rgb]{0.73,0.13,0.13}{##1}}}
\expandafter\def\csname PYG@tok@s2\endcsname{\def\PYG@tc##1{\textcolor[rgb]{0.73,0.13,0.13}{##1}}}
\expandafter\def\csname PYG@tok@sh\endcsname{\def\PYG@tc##1{\textcolor[rgb]{0.73,0.13,0.13}{##1}}}
\expandafter\def\csname PYG@tok@s1\endcsname{\def\PYG@tc##1{\textcolor[rgb]{0.73,0.13,0.13}{##1}}}
\expandafter\def\csname PYG@tok@mb\endcsname{\def\PYG@tc##1{\textcolor[rgb]{0.40,0.40,0.40}{##1}}}
\expandafter\def\csname PYG@tok@mf\endcsname{\def\PYG@tc##1{\textcolor[rgb]{0.40,0.40,0.40}{##1}}}
\expandafter\def\csname PYG@tok@mh\endcsname{\def\PYG@tc##1{\textcolor[rgb]{0.40,0.40,0.40}{##1}}}
\expandafter\def\csname PYG@tok@mi\endcsname{\def\PYG@tc##1{\textcolor[rgb]{0.40,0.40,0.40}{##1}}}
\expandafter\def\csname PYG@tok@il\endcsname{\def\PYG@tc##1{\textcolor[rgb]{0.40,0.40,0.40}{##1}}}
\expandafter\def\csname PYG@tok@mo\endcsname{\def\PYG@tc##1{\textcolor[rgb]{0.40,0.40,0.40}{##1}}}
\expandafter\def\csname PYG@tok@ch\endcsname{\let\PYG@it=\textit\def\PYG@tc##1{\textcolor[rgb]{0.25,0.50,0.50}{##1}}}
\expandafter\def\csname PYG@tok@cm\endcsname{\let\PYG@it=\textit\def\PYG@tc##1{\textcolor[rgb]{0.25,0.50,0.50}{##1}}}
\expandafter\def\csname PYG@tok@cpf\endcsname{\let\PYG@it=\textit\def\PYG@tc##1{\textcolor[rgb]{0.25,0.50,0.50}{##1}}}
\expandafter\def\csname PYG@tok@c1\endcsname{\let\PYG@it=\textit\def\PYG@tc##1{\textcolor[rgb]{0.25,0.50,0.50}{##1}}}
\expandafter\def\csname PYG@tok@cs\endcsname{\let\PYG@it=\textit\def\PYG@tc##1{\textcolor[rgb]{0.25,0.50,0.50}{##1}}}

\def\PYGZus{\char`\_}

\def\PYGZsh{\char`\#}

\def\PYGZhy{\char`\-}
\def\PYGZsq{\char`\'}
\def\PYGZdq{\char`\"}

\makeatother

\title{PhoNLP: A joint multi-task learning model for Vietnamese part-of-speech tagging, named entity recognition and dependency parsing}

\author{Linh The Nguyen \and Dat Quoc Nguyen\\
         VinAI Research, Hanoi, Vietnam \\
         \tt{\normalsize \{v.linhnt140, v.datnq9\}@vinai.io}}

\begin{document}
\maketitle

\begin{abstract}
We present the first multi-task learning model---named PhoNLP---for joint Vietnamese part-of-speech (POS) tagging, named entity recognition (NER) and dependency
parsing. Experiments on Vietnamese benchmark datasets show that PhoNLP produces state-of-the-art  results, outperforming a single-task learning approach that fine-tunes the pre-trained Vietnamese language model PhoBERT \cite{phobert} for each task independently.  We publicly release PhoNLP as an open-source toolkit under the Apache License 2.0. 
Although we specify PhoNLP for Vietnamese, our PhoNLP training and evaluation command scripts in fact can directly work for other languages that have a pre-trained BERT-based language model and gold annotated corpora available for the three tasks of POS tagging, NER and dependency parsing. 
We hope that PhoNLP can serve as a strong baseline and useful toolkit for future NLP research and applications to not only  Vietnamese but also the other languages. Our PhoNLP is available  at \url{https://github.com/VinAIResearch/PhoNLP}.
\end{abstract}

\vspace{-5pt}

\begin{figure*}[!t]
\centering
\includegraphics[width=12.5cm]{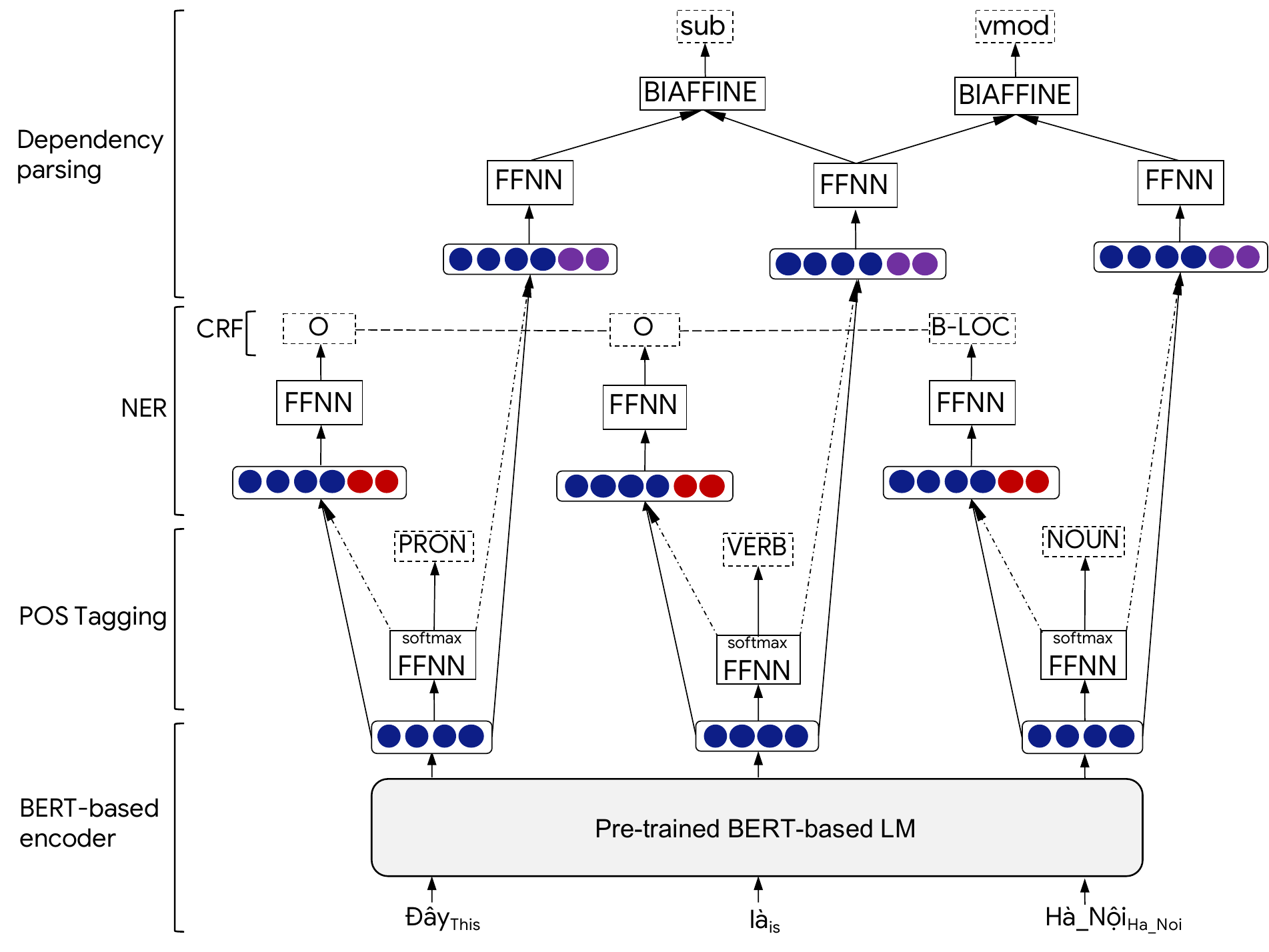}
{\small
\begin{tabular}{crllll}
\hline 
\textbf{ID} & \textbf{Form} & \textbf{POS} & \textbf{NER} & \textbf{Head} & \textbf{DepRel} \\
\hline
1 & Đây\textsubscript{This} & PRON & O & 2 & sub  \\
2 & là\textsubscript{is} & VERB& O & 0 & root \\
3 & Hà\_Nội\textsubscript{Ha\_Noi} & NOUN & B-LOC & 2 & vmod \\
\hline
\end{tabular}
}
\caption{Illustration of our PhoNLP model.}
\label{fig:architecture}
\end{figure*}

\section{Introduction}

Vietnamese NLP research has been significantly explored recently. It has been boosted by the success of the national project on Vietnamese language and speech processing (VLSP) KC01.01/2006-2010 and VLSP workshops that have run shared tasks since 2013.\footnote{\url{https://vlsp.org.vn/}} Fundamental tasks of POS tagging, NER and dependency parsing thus play important roles, providing useful features for many downstream application tasks such as machine translation \cite{7800281}, sentiment analysis \cite{BANG20182016IIP0038},  relation extraction \cite{9287471}, semantic parsing \cite{vitext2sql}, open information extraction \cite{3155133.3155171} and question answering \cite{NguyenNP_SWJ,3184558.3191535}. 
Thus, there is a need to develop  NLP toolkits for linguistic annotations w.r.t. Vietnamese POS tagging, NER and dependency parsing. 

VnCoreNLP \cite{vu-etal-2018-vncorenlp} is the previous public toolkit employing traditional feature-based machine learning models to handle those Vietnamese NLP tasks. However, VnCoreNLP is now no longer considered state-of-the-art  because its performance results are significantly outperformed by ones obtained when fine-tuning PhoBERT---the current state-of-the-art monolingual pre-trained language model for Vietnamese \cite{phobert}. Note that there are no publicly available fine-tuned BERT-based models for the three Vietnamese tasks. Assuming that there would be, a potential drawback might be that an NLP package wrapping such fine-tuned BERT-based models would take a large storage space, i.e. three times larger than the storage space used by  a BERT model \cite{devlin-etal-2019-bert}, thus  it would not be suitable for practical applications that require a smaller storage space. Jointly multi-task learning is a promising solution as it might help reduce the storage space. In addition, POS tagging, NER and dependency parsing are related tasks: POS tags are essential input features used for dependency parsing and POS tags are also used as additional features for NER.   Jointly multi-task learning thus might also help improve the performance results against the single-task learning \cite{Ruder2019Neural}.

In this paper, we present a new multi-task learning model---named PhoNLP---for joint POS tagging, NER and dependency parsing. In particular, given an input  sentence of words to PhoNLP, an encoding layer generates contextualized word embeddings that represent the input words. These contextualized word embeddings are fed into a POS tagging layer that is in fact a linear prediction layer \cite{devlin-etal-2019-bert} to predict POS tags for the corresponding input words. Each predicted POS tag is then represented by two ``soft'' embeddings that are later fed into NER and dependency parsing layers separately. 
More specifically, based on both the contextualized word embeddings and the ``soft'' POS tag embeddings, the NER layer uses a linear-chain CRF predictor \cite{Lafferty:2001} to predict NER labels for the  input words, while the dependency parsing layer uses a Biaffine classifier \cite{DozatM17} to predict dependency arcs between the words and another Biaffine classifier to label the predicted arcs. 
Our contributions are summarized as follows:


\begin{itemize}[leftmargin=*]
\setlength\itemsep{-1pt}
    \item To the best of our knowledge, PhoNLP is the first proposed model to jointly learn POS tagging, NER and dependency parsing for Vietnamese. 
    \item We discuss a data leakage issue in the Vietnamese benchmark datasets, that has not yet  been pointed out before. Experiments show that PhoNLP obtains state-of-the-art performance results, outperforming the PhoBERT-based single task learning.
    \item We publicly release PhoNLP as an open-source toolkit that is simple to setup and efficiently run from both the  command-line and Python API. We hope that PhoNLP can serve as a strong baseline and useful toolkit for future NLP research and downstream applications.
\end{itemize}

\section{Model description}

Figure \ref{fig:architecture} illustrates our PhoNLP architecture that can be viewed as a mixture of a BERT-based encoding layer and three decoding layers of POS tagging, NER and dependency parsing. 

\subsection{Encoder \& Contextualized embeddings}

Given an input sentence consisting of $n$ word tokens $w_1, w_2, ..., w_n$, the encoding layer employs PhoBERT to generate contextualized latent feature embeddings $\mathbf{e}_{i}$ each representing the $i^{th}$ word $w_i$: 

\begin{equation}
\mathbf{e}_{i} = \mathrm{PhoBERT\textsubscript{base}}\big({w}_{1:n}, i\big)
\end{equation}

In particular, the encoding layer employs the \textbf{PhoBERT\textsubscript{base}} version. Because PhoBERT uses BPE \cite{sennrich-etal-2016-neural} to segment the input sentence with subword units, the encoding layer in fact represents the $i^{th}$ word $w_i$ by using the contextualized embedding of its first subword.

\subsection{POS tagging}\label{ssec:pos}

Following a common manner when fine-tuning a pre-trained language model for a sequence labeling task \cite{devlin-etal-2019-bert}, the POS tagging layer is a linear prediction layer that is appended on top of the encoder. In particular, the POS tagging layer feeds the contextualized word embeddings $\mathbf{e}_{i}$ into a   feed-forward  network (FFNN\textsubscript{POS}) followed by a $\mathsf{softmax}$ predictor for POS tag prediction:

\begin{equation}
\mathbf{p}_{i} = \mathsf{softmax}\big(\mathrm{FFNN\textsubscript{POS}}\big(\mathbf{e}_{i}\big)\big) \label{eq2}
\end{equation}

\noindent where the output layer size of FFNN\textsubscript{POS}   is the number of POS tags. Based on probability vectors $\mathbf{p}_{i}$, a cross-entropy objective loss \textbf{$\mathcal{L}_{\text{POS}}$} is calculated for POS tagging during training.

\subsection{NER}\label{ssec:ner}

The NER layer creates a sequence of vectors $\mathbf{v}_{1:n}$ in which each $\mathbf{v}_{i}$ is resulted in  by concatenating the contextualized word embedding $\mathbf{e}_{i}$ and a ``soft'' POS tag embedding $\mathbf{t}_{i}^{(1)}$:

\begin{equation}
\mathbf{v}_{i} = \mathbf{e}_{i}  \circ   \mathbf{t}_{i}^{(1)}
\label{equa:ner}
\end{equation}

\noindent where following \newcite{hashimoto-etal-2017-joint}, the ``soft'' POS tag embedding $\mathbf{t}_{i}^{(1)}$ is computed by multiplying a label weight matrix $\mathbf{W}^{(1)}$ with the corresponding probability vector $\mathbf{p}_{i}$:

\begin{equation*}
 \mathbf{t}_{i}^{(1)} = \mathbf{W}^{(1)}\mathbf{p}_{i} 
\end{equation*}

The NER layer then passes each vector   $\mathbf{v}_{i}$ into a  FFNN (FFNN\textsubscript{NER}):

\begin{equation}
 \mathbf{h}_{i} = \mathrm{FFNN\textsubscript{NER}}\big(\mathbf{v}_{i}\big) \label{eq4}
\end{equation}

\noindent where  the output layer size of FFNN\textsubscript{NER} is the number of BIO-based NER labels.

The NER layer  feeds the output vectors $\mathbf{h}_{i}$ into a linear-chain
CRF predictor for NER label prediction \cite{Lafferty:2001}. A cross-entropy loss \textbf{$\mathcal{L}_{\text{NER}}$} is calculated for NER during
training while the Viterbi algorithm is used for inference.

\subsection{Dependency parsing}

The dependency parsing layer creates vectors $\mathbf{z}_{1:n}$ in which each $\mathbf{z}_{i}$ is resulted in by concatenating $\mathbf{e}_{i}$ and another ``soft'' POS tag embedding $\mathbf{t}_{i}^{(2)}$:

\begin{eqnarray}
\mathbf{z}_{i} &=& \mathbf{e}_{i}  \circ   \mathbf{t}_{i}^{(2)} \label{equa:posdep}  \\
\mathbf{t}_{i}^{(2)} &=& \mathbf{W}^{(2)}\mathbf{p}_{i} \nonumber
\end{eqnarray}

Following \newcite{DozatM17}, the dependency parsing layer uses FFNNs to split $\mathbf{z}_{i}$  into \emph{head} and \emph{dependent} representations:

 \begin{eqnarray}
\mathbf{h}_{i}^{(\textsc{a-h})}  &=& \mathrm{FFNN}_{\text{Arc-Head}}\big(\mathbf{z}_{i}\big)  \label{equa:fc6}  \\ 
\mathbf{h}_{i}^{(\textsc{a-d})}  &=& \mathrm{FFNN}_{\text{Arc-Dep}}\big(\mathbf{z}_{i}\big)  \\ 
\mathbf{h}_{i}^{(\textsc{l-h})}  &=& \mathrm{FFNN}_{\text{Label-Head}}\big(\mathbf{z}_{i}\big)  \\ 
\mathbf{h}_{i}^{(\textsc{l-d})}  &=& \mathrm{FFNN}_{\text{Label-Dep}}\big(\mathbf{z}_{i}\big)  \label{equa:fc9} 
\end{eqnarray}

To predict potential dependency arcs, based on input vectors $\mathbf{h}_{i}^{(\textsc{a-h})}$ and $\mathbf{h}_{j}^{(\textsc{a-d})}$, the parsing layer uses  a Biaffine classifier's variant \cite{qi-etal-2018-universal} that additionally takes into account the distance and relative ordering between two words to produce a probability distribution of
arc heads for each word. 
For inference, the Chu–Liu/Edmonds' algorithm is used to find a maximum spanning tree \cite{chuliu,Edmonds}. 
The parsing layer also uses another Biaffine classifier to label the predicted arcs,  based on input vectors $\mathbf{h}_{i}^{(\textsc{l-h})}$ and $\mathbf{h}_{j}^{(\textsc{l-d})}$. An objective loss \textbf{$\mathcal{L}_{\text{DEP}}$} is computed by summing a cross entropy loss for unlabeled dependency parsing and another cross entropy loss  for dependency label prediction during training based on gold arcs and arc labels.

%
%
%

\subsection{Joint multi-task learning}

The final training objective loss \textbf{$\mathcal{L}$} of our model PhoNLP is the weighted sum of the POS tagging loss {$\mathcal{L}_{\text{POS}}$}, the NER loss {$\mathcal{L}_{\text{NER}}$} and the dependency parsing loss {$\mathcal{L}_{\text{DEP}}$}:

\begin{equation}
\textbf{$\mathcal{L}$} = \lambda_1\mathcal{L}_{\text{POS}} + \lambda_2\mathcal{L}_{\text{NER}} + (1 - \lambda_1 - \lambda_2)\mathcal{L}_{\text{DEP}}
\end{equation}

\paragraph{Discussion:} Our PhoNLP can be viewed as an extension of previous joint POS tagging and dependency parsing models \cite{hashimoto-etal-2017-joint,li-etal-2018-joint-learning,nguyen-verspoor-2018-improved,NguyenALTA2019,kondratyuk-straka-2019-75}, where we additionally incorporate a CRF-based prediction layer for NER. Unlike \newcite{hashimoto-etal-2017-joint}, \newcite{nguyen-verspoor-2018-improved}, \newcite{li-etal-2018-joint-learning} and \newcite{NguyenALTA2019} that use BiLSTM-based encoders to extract contextualized feature embeddings, we use a BERT-based encoder. \newcite{kondratyuk-straka-2019-75} also employ a BERT-based encoder. However, different from PhoNLP where we construct a hierarchical architecture over the POS tagging and dependency parsing layers, \newcite{kondratyuk-straka-2019-75} do not make use of POS tag embeddings for dependency parsing.\footnote{In our preliminary experiments, not feeding the POS tag embeddings into the dependency parsing layer decreases the performance.}

\section{Experiments}

\subsection{Setup}

\subsubsection{Datasets}

To conduct experiments, we use the  benchmark datasets of the VLSP 2013 POS tagging   dataset,\footnote{\url{https://vlsp.org.vn/vlsp2013/eval}} the VLSP 2016 NER  dataset \cite{JCC13161} and  the VnDT dependency treebank v1.1 \newcite{Nguyen2014NLDB}, following the setup used by the VnCoreNLP toolkit  \cite{vu-etal-2018-vncorenlp}. Here, VnDT is converted from the Vietnamese constituent treebank \cite{nguyen-etal-2009-building}.

\paragraph{Data leakage issue:} We further discover an issue of data leakage, that has not yet been pointed out before. That is, all sentences from the VLSP 2016 NER dataset and the VnDT treebank are included in the VLSP 2013 POS tagging dataset. In particular, 90+\% of sentences from both validation and test sets for NER and dependency parsing are included in the POS tagging training set, resulting in an unrealistic evaluation scenario where the POS tags are used as input features for NER and dependency parsing.

To handle the data leakage issue, we have to re-split the VLSP 2013 POS tagging dataset to avoid the data leakage issue: The POS tagging validation/test set now only contains sentences that appear in the union of the NER and dependency parsing validation/test sets (i.e. the validation/test sentences for NER and dependency parsing only appear in the POS tagging validation/test set).  
In addition, there are 594 duplicated sentences in the VLSP 2013 POS tagging dataset (here, sentence duplication is not found in the union of the NER and dependency parsing sentences). Thus we have to  perform duplication removal on the POS tagging dataset. 
Table \ref{tab:Datasets} details the statistics of the experimental datasets. 

\begin{table}[!t]
\centering
\resizebox{7.5cm}{!}{
\begin{tabular}{l|l|l|l}
\hline
\textbf{Task} & \textbf{\#train} & \textbf{\#valid} & \textbf{\#test} \\ 
\hline 
{POS tagging (leakage)} & {27000} & {870} & {2120} \\
\hdashline
POS tagging (re-split) & 23906 & 2009 & 3481\\ 
\hline 
NER  & 14861 & 2000 & 2831 \\
\hline 
Dependency parsing & 8977 & 200 & 1020 \\
\hline
\end{tabular}    
}
\caption{Dataset statistics. \textbf{\#train}, \textbf{\#valid} and \textbf{\#test} denote the numbers of training, validation and test sentences, respectively. Here, 
``{POS tagging (leakage)}'' and ``POS tagging (re-split)'' refer to the statistics for  POS tagging before and after re-splitting \& sentence duplication removal, respectively.}
\label{tab:Datasets}
\end{table}

\subsubsection{Implementation}

PhoNLP is implemented based on PyTorch \cite{NEURIPS2019_9015}, employing the PhoBERT encoder implementation available from the $\mathrm{transformers}$ library \cite{wolf-etal-2020-transformers} and the Biaffine classifier implementation from \newcite{qi-etal-2020-stanza}. We set both the label weight matrices $\mathbf{W}^{(1)}$ and $\mathbf{W}^{(2)}$ to have 100 rows, resulting in 100-dimensional soft POS tag embeddings. In addition, following \newcite{qi-etal-2018-universal,qi-etal-2020-stanza}, FFNNs in equations \ref{equa:fc6}--\ref{equa:fc9} use 400-dimensional output layers. 

We use the AdamW optimizer \cite{loshchilov2018decoupled} and a fixed batch size at 32, and train for 40 epochs. The sizes of training sets are different, in which the POS tagging
training set is the largest, consisting of 23906 sentences. Thus for each training epoch, we repeatedly sample from the NER and dependency parsing training sets to fill the gaps between the training set sizes. We perform a grid search to select the initial AdamW learning rate, $\lambda_1$ and $\lambda_2$. We find the optimal initial AdamW learning rate, $\lambda_1$ and $\lambda_2$ at 1e-5, 0.4 and 0.2, respectively. Here, we compute the average of the POS tagging accuracy, NER F\textsubscript{1}-score and   dependency parsing score LAS after each training epoch on the validation sets. We select the model checkpoint that produces the highest average score over the validation sets to apply to the test sets.  Each of our reported scores is an average over 5 runs with different random seeds. 



\subsection{Results}


Table \ref{tab:results} presents results obtained for our PhoNLP and compares them with those of a baseline approach of single-task training. For the  single-task training approach: (i) We follow a common approach to fine-tune a pre-trained language model for POS tagging, appending a linear prediction layer on top of PhoBERT,  as briefly described in Section \ref{ssec:pos}. (ii) For NER,  instead of a linear prediction layer, we append a CRF prediction layer on top of PhoBERT. (iii) For dependency parsing, predicted POS tags are produced by the learned single-task POS tagging model; then POS tags are represented by embeddings that are concatenated with the corresponding PhoBERT-based contextualized word embeddings, resulting in a sequence of input vectors for the Biaffine-based classifiers for dependency parsing \cite{qi-etal-2018-universal}. Here, the  single-task training approach is based on the PhoBERT\textsubscript{base} version,  employing the same hyper-parameter tuning and model selection strategy that we use for PhoNLP.

\begin{table}[!t]
\centering
\def\arraystretch{1.2}
\resizebox{7.5cm}{!}{
\begin{tabular}{ll|l|l|l|l}
\hline
& \textbf{Model} & \textbf{POS} & \textbf{NER} & \textbf{LAS} & \textbf{UAS} \\ 
\hline 
\multirow{2}{*}{\rotatebox[origin=c]{90}{{Leak.}}}& Single-task & 96.7$^\dagger$ & 93.69 & 78.77$^\dagger$ & 85.22$^\dagger$ \\
\cdashline{2-6}
& PhoNLP &  \textbf{96.76} & \textbf{94.41} & \textbf{79.11} & \textbf{85.47}\\
\hline
\hline
\multirow{2}{*}{\rotatebox[origin=c]{90}{{Re-spl}}}& Single-task & 93.68 & 93.69 & 77.89 & 84.78 \\
\cdashline{2-6}
& PhoNLP & \textbf{93.88}  & \textbf{94.51}  & \textbf{78.17}  & \textbf{84.95} \\
\hline
\end{tabular}     
}
\caption{Performance results (in \%) on the test sets  for POS tagging (i.e. accuracy), NER (i.e. F\textsubscript{1}-score) and dependency parsing (i.e. LAS and UAS scores). ``Leak.'' abbreviates ``leakage'', denoting the results obtained w.r.t. the data leakage issue. ``Re-spl'' denotes the results obtained w.r.t.  the data re-split and duplication removal for POS tagging to avoid the data leakage issue. ``Single-task'' refers to as the single-task training approach. 
$\dagger$ denotes scores taken from the PhoBERT paper \cite{phobert}. Note that ``Single-task'' NER is not affected by the data leakage issue.   
}
\label{tab:results}
\end{table}

Note that PhoBERT helps produce state-of-the-art results for multiple Vietnamese NLP tasks (including but not limited to POS tagging, NER and dependency parsing in a single-task training strategy), and obtains  higher performance results than VnCoreNLP.    
However, in both the PhoBERT and VnCoreNLP papers \cite{phobert,vu-etal-2018-vncorenlp}, results for POS tagging and dependency parsing are reported w.r.t. the data leakage issue. Our ``Single-task'' results in Table \ref{tab:results} regarding ``Re-spl'' (i.e. the data re-split and duplication removal for POS tagging to avoid the data leakage issue) can be viewed as new PhoBERT results for a proper experimental setup. Table \ref{tab:results}  shows that in both setups ``Leak.'' and ``Re-spl'',  our joint multi-task training approach PhoNLP performs better than the PhoBERT-based single-task training approach, thus resulting in state-of-the-art performances for the three tasks of Vietnamese POS tagging, NER and dependency parsing.

\section{PhoNLP toolkit}

We present in this section a basic usage of our PhoNLP toolkit. 
We make PhoNLP simple to setup, i.e.  users can install PhoNLP from either source or $\mathsf{pip}$ (e.g. $\mathsf{pip3\ install\ phonlp}$). We also aim to make PhoNLP simple to run from both the command-line and the Python API. For example, annotating a corpus with POS tagging, NER and dependency parsing can be performed by using a simple command as in Figure \ref{fig:command}.

Assume that the input file ``{\ttfamily input.txt}'' in Figure \ref{fig:command} contains a sentence ``Tôi đang làm\_việc tại VinAI .'' (I\textsubscript{Tôi} am\textsubscript{đang} working\textsubscript{làm\_việc} at\textsubscript{tại} VinAI). Table \ref{tab:format} shows the annotated output  in plain text form for this
sentence. Similarly, we also get the same output by using the Python API as simple as in Figure \ref{fig:code}.  
Furthermore,  commands to (re-)train  and evaluate PhoNLP using gold annotated corpora are detailed in the PhoNLP GitHub repository. Note that it is absolutely possible to directly employ our PhoNLP (re-)training and evaluation command scripts for other languages that have gold annotated corpora available for the three tasks and a pre-trained BERT-based language model available from the $\mathrm{transformers}$ library. 

\begin{figure}[!t]
{\ttfamily python3 run\_phonlp.py {-}{-}save\_dir ./pretrained\_phonlp {-}{-}mode \\ annotate {-}{-}input\_file input.txt {-}{-}output\_file output.txt}
\caption{Minimal command to run PhoNLP. Here ``save\_dir'' denote the path to the local machine folder that stores the pre-trained PhoNLP model.}
\label{fig:command}
\end{figure}

\begin{table}[!t]
\centering
\begin{tabular}{llllll}
1 & Tôi & P & O & 3 & sub \\
2 & đang & R & O & 3 & adv \\ 
3 & làm\_việc & V & O & 0 & root \\
4 & tại & E & O & 3 & loc \\
5 & VinAI & Np & B-ORG & 4 & pob \\
6 & . & CH & O & 3 & punct \\
\end{tabular}    
\caption{The output in the output file ``{\ttfamily output.txt}'' for the sentence ``Tôi đang làm\_việc tại VinAI .'' from the input file ``{\ttfamily input.txt}'' in Figure \ref{fig:command}. The output is formatted with 6 columns representing word index, word form, POS tag, NER label, head index of the current word and its dependency relation type.}
\label{tab:format}
\end{table}

\paragraph{Speed test:} We perform a sole CPU-based speed test using a personal computer with Intel Core i5 8265U 1.6GHz \& 8GB of memory. For a GPU-based speed test, we employ a machine with a single NVIDIA RTX 2080Ti GPU. For performing the three NLP tasks jointly, PhoNLP obtains a  speed at {15 sentences per second} for the CPU-based test and {129 sentences per second} for the GPU-based test, respectively, with an average of 23 word tokens per sentence and a batch size of 8. 

\begin{figure*}[!t]
\begin{Verbatim}[commandchars=\\\{\}]
\PYG{k+kn}{import} \PYG{n+nn}{phonlp}
\PYG{c+c1}{\PYGZsh{} Automatically download the pre\PYGZhy{}trained PhoNLP model}
\PYG{c+c1}{\PYGZsh{} and save it in a local machine folder}
\PYG{n}{phonlp}\PYG{o}{.}\PYG{n}{download}\PYG{p}{(}\PYG{n}{save\PYGZus{}dir}\PYG{o}{=}\PYG{l+s+s1}{\PYGZsq{}./pretrained\PYGZus{}phonlp\PYGZsq{}}\PYG{p}{)}
\PYG{c+c1}{\PYGZsh{} Load the pre\PYGZhy{}trained PhoNLP model}
\PYG{n}{model} \PYG{o}{=} \PYG{n}{phonlp}\PYG{o}{.}\PYG{n}{load}\PYG{p}{(}\PYG{n}{save\PYGZus{}dir}\PYG{o}{=}\PYG{l+s+s1}{\PYGZsq{}./pretrained\PYGZus{}phonlp\PYGZsq{}}\PYG{p}{)}
\PYG{c+c1}{\PYGZsh{} Annotate a corpus}
\PYG{n}{model}\PYG{o}{.}\PYG{n}{annotate}\PYG{p}{(}\PYG{n}{input\PYGZus{}file}\PYG{o}{=}\PYG{l+s+s1}{\PYGZsq{}input.txt\PYGZsq{}}\PYG{p}{,} \PYG{n}{output\PYGZus{}file}\PYG{o}{=}\PYG{l+s+s1}{\PYGZsq{}output.txt\PYGZsq{}}\PYG{p}{)}
\PYG{c+c1}{\PYGZsh{} Annotate a sentence}
\PYG{n}{model}\PYG{o}{.}\PYG{n}{print\PYGZus{}out}\PYG{p}{(}\PYG{n}{model}\PYG{o}{.}\PYG{n}{annotate}\PYG{p}{(}\PYG{n}{text}\PYG{o}{=}\PYG{l+s+s2}{\PYGZdq{}Tôi đang làm\PYGZus{}việc tại VinAI .\PYGZdq{}}\PYG{p}{))}
\end{Verbatim}
\caption{A simple and complete example code for using PhoNLP in Python.}
\label{fig:code}
\end{figure*}

\section{Conclusion and future work}

We have presented the first multi-task learning model PhoNLP for joint  POS tagging, NER and dependency parsing in Vietnamese. Experiments on Vietnamese benchmark datasets show that PhoNLP outperforms its strong fine-tuned PhoBERT-based single-task training baseline, producing state-of-the-art performance results. We publicly release PhoNLP as an easy-to-use open-source toolkit and hope that PhoNLP can facilitate future NLP research and applications. 
In future work, we will also apply  PhoNLP  to other languages.


\bibliography{refs}
\bibliographystyle{acl_natbib}

\end{document}